\documentclass[letterpaper, 10pt, conference]{ieeeconf}
\IEEEoverridecommandlockouts
\usepackage{newtxtext,newtxmath}
\usepackage{graphicx}
\usepackage{amsmath}
\usepackage{booktabs}
\usepackage[table]{xcolor}
\usepackage[hidelinks]{hyperref}
\usepackage{cite}
\usepackage{xspace}
\usepackage{balance}
\usepackage{placeins}

\definecolor{TableBand}{HTML}{F2F4F5}


\newcommand{\methodA}{\textsc{Online}}
\newcommand{\methodG}{\textsc{FixG}}
\newcommand{\methodB}{\textsc{FixBa}}
\newcommand{\methodGB}{\textsc{FixG+Ba}}
\newcommand{\lsFGBA}{\textsc{FG--BA}}
\newcommand{\lsFGBZ}{\textsc{FG--B0}}
\newcommand{\lsGEBA}{\textsc{GE--BA}}
\newcommand{\lsGEBZ}{\textsc{GE--B0}}
\newcommand{\NumCoreSeq}{12\xspace}
\newcommand{\NumCoreRuns}{48\xspace}

\newcommand{\NumRuntimePairs}{3\xspace}

\newcommand{\NumRuntimeTimedScans}{3144\xspace}
\newcommand{\NumRuntimeWarmupScans}{100\xspace}

\newcommand{\NumDynamicInitPhases}{3\xspace}

\newcommand{\NumFastDynamicInitPairs}{4\xspace}

\newcommand{\NumDynamicFactorialPhases}{12\xspace}
\newcommand{\NumDynamicFactorialRuns}{48\xspace}

\newcommand{\NumCoupledStateRuns}{10\xspace}

\newcommand{\CoupledStateTiltDeg}{2\xspace}
\newcommand{\CoupledStateInitialBA}{0.342\,m/s$^2$\xspace}
\newcommand{\CoupledStateInitialHMax}{$2.0\times10^{-7}$\,m/s$^2$\xspace}
\newcommand{\CoupledHhsGRange}{0.73--0.74$^\circ$\xspace}
\newcommand{\CoupledHhsBARange}{0.122--0.126\,m/s$^2$\xspace}
\newcommand{\CoupledHhsHRange}{0.0077--0.0091\,m/s$^2$\xspace}
\newcommand{\CoupledHhsPoseRange}{4.2--8.4\,mm\xspace}

\newcommand{\NumDynamicFixedBaFailureOutcomes}{2\xspace}
\newcommand{\NumStrongFixBaATEWorse}{4\xspace}

\newcommand{\StrongFixBaATERange}{+1.64\% to +65.82\%\xspace}

\newcommand{\FastDynamicRescueOnlineZ}{0.625\,m\xspace}
\newcommand{\FastDynamicRescueFixGZ}{0.170\,m\xspace}
\newcommand{\FastDynamicRescueOnlineATE}{0.905\,m\xspace}
\newcommand{\FastDynamicRescueFixGATE}{5.074\,m\xspace}

\newcommand{\RuntimeAlgebraReduction}{7.5\%\xspace}
\newcommand{\RuntimeAlgebraCoreShare}{0.83\%\xspace}
\newcommand{\RuntimeMatchingCoreShare}{88.2\%\xspace}
\newcommand{\RuntimeAttributableCoreSaving}{0.06\%\xspace}

\newcommand{\RuntimeCoreMedianChangeRange}{-11.6\% to +3.1\%\xspace}

\newcommand{\NumBoundaryRepeats}{3\xspace}
\newcommand{\GravWanderMax}{4.0$^\circ$\xspace}
\newcommand{\MinPath}{35\,m\xspace}
\newcommand{\MaxPath}{3.2\,km\xspace}
\newcommand{\AlignSec}{10\,s\xspace}
\newcommand{\AlignM}{30\,m\xspace}
\newcommand{\GZMedian}{+0.9\%\xspace}
\newcommand{\GZAbsMedian}{1.1\%\xspace}
\newcommand{\GZRange}{-6.7--+2.6\%\xspace}
\newcommand{\GATEMedian}{-0.1\%\xspace}
\newcommand{\GATEAbsMedian}{0.9\%\xspace}
\newcommand{\GATERange}{-4.0--+5.4\%\xspace}
\newcommand{\EquivMargin}{5\%\xspace}
\newcommand{\StrictEquivMargin}{2\%\xspace}

\newcommand{\EquivPBoth}{$<10^{-4}$\xspace}
\newcommand{\StrictEquivPBoth}{$0.020$\xspace}
\newcommand{\EquivCIZ}{-0.11\% [-1.54, +1.34]\%\xspace}
\newcommand{\EquivCIATE}{+0.09\% [-1.16, +1.35]\%\xspace}
\newcommand{\FixGAbsDeltaZMedian}{0.017\,m\xspace}
\newcommand{\FixGAbsDeltaATEMedian}{0.014\,m\xspace}
\newcommand{\FamilyEquivPBoth}{$0.002$\xspace}

\newcommand{\ReproEquivCIZ}{-0.28\% [-1.84, +1.30]\%\xspace}
\newcommand{\ReproEquivCIATE}{-0.11\% [-1.50, +1.31]\%\xspace}

\newcommand{\ReproStrictEquivPBoth}{$0.040$\xspace}

\newcommand{\HallFixGZ}{+2.3\%\xspace}
\newcommand{\HallFixGZDeltaMm}{+1.0\,mm\xspace}
\newcommand{\HallFixGATE}{-0.45\%\xspace}
\newcommand{\HallFixGATEDeltaMm}{-5.7\,mm\xspace}
\newcommand{\NumLioSamSeq}{4\xspace}

\newcommand{\LioSamGZRange}{-0.37--+0.46\%\xspace}
\newcommand{\LioSamGATERange}{-0.07--+0.18\%\xspace}
\newcommand{\LioSamBothZRange}{-3.20--+0.76\%\xspace}
\newcommand{\LioSamBothATERange}{-1.09--+0.27\%\xspace}

\newcommand{\DirStrongFGZRange}{+1.0\% to +1.2\%\xspace}
\newcommand{\DirMidFGZRange}{-1.7\% to -1.2\%\xspace}
\newcommand{\DirMidFGATERange}{-0.75\% to +0.03\%\xspace}
\newcommand{\DirGEATERange}{-0.07\% to +0.04\%\xspace}

\newcommand{\NumDirRepeatsPerCell}{3\xspace}
\newcommand{\NumHallDirBeneficialCells}{6\xspace}
\newcommand{\NumHallDirResetOutcomes}{1\xspace}
\newcommand{\TuhhDirFGStrongZ}{-64.6\%\xspace}
\newcommand{\TuhhDirFGStrongATE}{+97.6\%\xspace}
\newcommand{\TuhhDirGEZTwo}{+450.0\%\xspace}
\newcommand{\TuhhDirGEATETwo}{+108.4\%\xspace}
\newcommand{\NumTuhhDirArcWarnings}{2\xspace}

\newcommand{\BAMedianZ}{+1.5\%\xspace}
\newcommand{\BothMedianZ}{+1.9\%\xspace}
\newcommand{\ProxyBAZ}{+7.3~pp\xspace}
\newcommand{\InteractionZ}{-0.8~pp\xspace}
\newcommand{\InteractionPZ}{$0.470$\xspace}
\newcommand{\InteractionATE}{-0.8~pp\xspace}
\newcommand{\InteractionPATE}{$0.151$\xspace}

\newcommand{\DropTwoGZ}{-11.2\%\xspace}
\newcommand{\DropTwoGATE}{+16.8\%\xspace}
\newcommand{\DropThreeGZ}{+93.1\%\xspace}
\newcommand{\DropThreeGATE}{+119.9\%\xspace}

\newcommand{\RangeGZ}{+0.2\%\xspace}
\newcommand{\RangeGATE}{-4.0\%\xspace}
\newcommand{\FovGZ}{+6.9\%\xspace}
\newcommand{\FovGATE}{+3.5\%\xspace}

\newcommand{\NumImuStressSeq}{2\xspace}

\newcommand{\ImuNoiseMin}{0.01\xspace}
\newcommand{\ImuNoiseMax}{10\xspace}
\newcommand{\ImuDayZMin}{-10.7\%\xspace}
\newcommand{\ImuDayZMax}{-5.0\%\xspace}
\newcommand{\ImuNightZMin}{-3.2\%\xspace}
\newcommand{\ImuNightZMax}{+1.7\%\xspace}
\newcommand{\ImuDayATEMin}{-1.2\%\xspace}
\newcommand{\ImuDayATEMax}{-0.2\%\xspace}
\newcommand{\ImuNightATEMin}{+2.3\%\xspace}
\newcommand{\ImuNightATEMax}{+9.0\%\xspace}
\newcommand{\ImuStressAttMax}{0.052$^\circ$\xspace}
\newcommand{\NumInitStressRuns}{12\xspace}

\newcommand{\InitStressAngles}{$0.5^\circ/1^\circ/2^\circ$\xspace}
\newcommand{\InitErrZMin}{-3.5\%\xspace}
\newcommand{\InitErrZMax}{+2.1\%\xspace}
\newcommand{\InitErrATEMin}{-2.3\%\xspace}
\newcommand{\InitErrATEMax}{+1.8\%\xspace}
\newcommand{\InitErrAttMax}{0.013$^\circ$\xspace}

\newcommand{\HhsDropThreeGZ}{+8.0\%\xspace}
\newcommand{\HhsDropThreeGATE}{+7.3\%\xspace}
\newcommand{\HhsDropFiveGZ}{+27.9\%\xspace}
\newcommand{\HhsDropFiveGATE}{+34.5\%\xspace}

\newcommand{\NumWarmRepeatPairs}{3\xspace}
\newcommand{\NumWarmPhases}{3\xspace}

\newcommand{\WarmVehCleanATE}{-0.83\%\xspace}

\newcommand{\WarmHhsCleanATE}{-0.30\%\xspace}

\newcommand{\WarmVehATEInteraction}{+4.16\,m\xspace}

\newcommand{\WarmHhsATEInteraction}{+1.01\,m\xspace}

\title{\LARGE\bfseries Does Online Gravity Estimation Matter?\\
Revisiting a Silent Design Split in LiDAR-Inertial Odometry}

\author{{\small Jie Xu \quad Ziyi Jin \quad Kangjin Yu \quad Can Jiang}\\[2pt]
{\small Hongjun Huang \quad Tongxing Jin \quad Hongkun Luo \quad Zhongpu Xia}\\[2pt]
{\small Anyverse Dynamics}}

\makeatletter
\def\ps@firstcontact{%
  \let\@oddhead\@empty\let\@evenhead\@empty
  \def\@oddfoot{\parbox[b]{\textwidth}{\footnotesize\raggedright
    Corresponding author: Zhongpu Xia.\quad
    Contact (Jie Xu): \texttt{jeff\_xu\_0503@foxmail.com}.\par
    \vspace{4pt}\centering\normalsize\thepage}}%
  \let\@evenfoot\@oddfoot}
\makeatother

\hypersetup{
  pdftitle={Does Online Gravity Estimation Matter? Revisiting a Silent Design Split in LiDAR-Inertial Odometry},
  pdfauthor={Jie Xu; Ziyi Jin; Kangjin Yu; Can Jiang; Hongjun Huang; Tongxing Jin; Hongkun Luo; Zhongpu Xia}
}

\begin{document}
\maketitle
\thispagestyle{firstcontact}
\pagestyle{plain}

\begin{abstract}
LiDAR--inertial odometry (LIO) systems differ in whether they continue estimating
gravity after initialization. We compare four gravity--bias state configurations
in each of FAST-LIO2 and LIO-SAM, then separately test a gravity-direction factor.
Across \NumCoreSeq{} dataset sequences evaluated with FAST-LIO2,
fixing gravity under continuous LiDAR correction produces mean paired changes
in vertical and 3D position errors with 90\% confidence intervals within $\pm$\StrictEquivMargin{}.
Tests on \NumLioSamSeq{} sequences with LIO-SAM likewise show no consistent benefit from
online gravity. Multi-second LiDAR outages, unlike reduced range or field of
view, reveal trajectory-dependent costs of fixing gravity. A history-matched
23D-to-21D switch places the repeatable 3D error increase after LiDAR updates
resume. Under 5-s outages, a direction factor from the same IMU used for
preintegration improves accuracy on Hall05 but worsens both errors with online
gravity on TUHH. Dynamic-start tests also show fixed-bias failures at
particular starting phases. We recommend keeping gravity and accelerometer
bias online for robustness; use a direction factor only after verifying
vertical and 3D accuracy gains under the intended operating conditions.

\end{abstract}

\newcommand{\EarlyTeaser}{}
\begin{figure}[!t]
\centering
\includegraphics[width=\columnwidth]{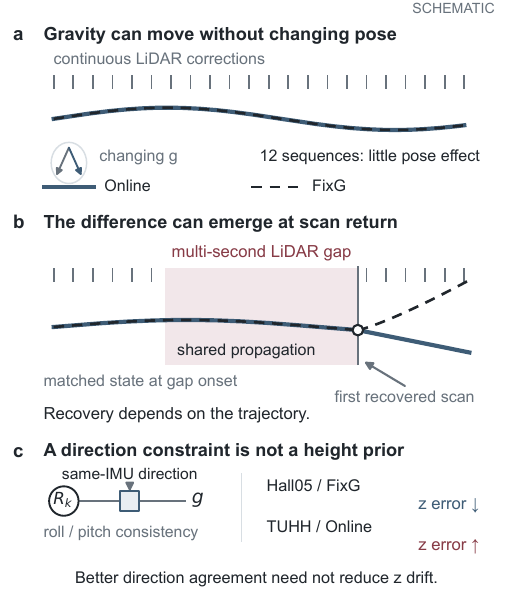}
\caption{Study logic (schematic, not measured paths). \textbf{a}, Continuous
LiDAR correction can mask gravity-state motion. \textbf{b}, Matched states
share propagation; scan return can reveal different recovery.
\textbf{c}, A direction factor from the same IMU constrains attitude, not height:
the tested dropout cases show opposite vertical-error responses.}
\label{fig:teaser}
\end{figure}

\noindent{\small Code, evidence, and video:\par
\url{https://github.com/jiejie567/rethink-lio-gravity}\par}

\section{Introduction}
\label{sec:intro}

LiDAR--inertial odometry (LIO) combines inertial propagation with corrections
from scan registration. This propagation requires gravity and accelerometer
bias, but systems differ in whether these quantities remain estimated states
after initialization. FAST-LIO2 and Point-LIO retain gravity direction on $S^2$ online
\cite{xu2022fastlio2,he2023pointlio}, whereas LIO-SAM fixes it
\cite{shan2020liosam}. Lightning-LM goes further, reducing its filter from
23D to 12D by removing online estimation of accelerometer bias $\mathbf b_a$,
gravity $\mathbf g$, and the extrinsics \cite{gao2026lightninglm}. These designs
raise a practical question: which states need to remain online, and under
what operating conditions?

Comparing published trajectory errors cannot answer this question. The systems
also differ in registration, maps, trajectory models, and parameter settings. Even within
one system, a changing gravity estimate does not establish that estimating it
improves pose. Under weak excitation, changes in gravity, attitude, and
accelerometer bias can explain similar inertial residuals. Gravity may then
wander while LiDAR corrections keep the trajectory accurate. Removing gravity
could suppress this behavior, but could also remove an adjustment needed after
poor initialization or a long interruption of LiDAR correction.

Gravity adds only two tangent coordinates, so removing it offers little
computational incentive if point-cloud matching dominates runtime. The stronger
reason to fix gravity would be fewer weakly observable states without losing
accuracy or recovery. Testing that argument requires conditions beyond nominal
operation, where frequent scans may conceal the consequences of either choice.

We test whether online gravity estimation remains necessary for accurate pose
estimation when LiDAR updates are frequently accepted. We then weaken the available geometry
or remove corrections altogether to determine whether these conditions have
the same effect. Dynamic startup tests a different source of error: motion
during the shared acceleration-mean initialization. Keeping $\mathbf g$ and
$\mathbf b_a$ online permits subsequent adjustment, although their individual
estimates need not become physically correct. The experiments distinguish
improved trajectory estimation from accurate identification of these two
coupled quantities.

\ifdefined\EarlyTeaser\else
\begin{figure}[!t]
\centering
\includegraphics[width=\columnwidth]{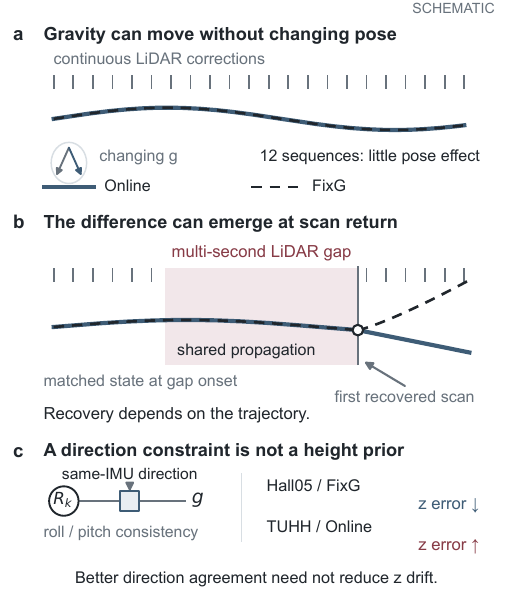}
\caption{Study logic (schematic, not measured paths). \textbf{a}, Continuous
LiDAR correction can mask gravity-state motion. \textbf{b}, Matched states
share propagation; scan return can reveal different recovery.
\textbf{c}, A direction factor from the same IMU constrains attitude, not height:
the tested dropout cases show opposite vertical-error responses.}
\label{fig:teaser}
\end{figure}

\fi

Adding a gravity-direction factor is a separate design choice. Fixing gravity
removes a state, whereas the factor adds a residual that constrains attitude
relative to an estimated down direction. It does not measure height. Any
vertical-error reduction must arise indirectly through attitude, inertial
propagation, or scan registration. In particular, a direction estimate obtained
from the same IMU used for preintegration is not an independent observation.
We refer to this as a \emph{same-IMU direction factor}. We test whether closer
direction agreement lowers trajectory error, and whether the answer depends
on factor weight or on whether gravity is estimated online.

Runtime correction nulling cannot test state removal. Suppressed coordinates
remain in the covariance and influence retained variables through
cross-correlation. We instead compile four FAST-LIO2 state configurations,
each with a manifold that retains or removes gravity and accelerometer bias.
The front end, data, initialization, calibration, and parameter settings remain fixed. A true
$2\times2$ LIO-SAM state ablation provides a descriptive cross-architecture
check. Within that graph, factor on/off and fixed/online gravity isolate the
same-IMU direction factor.

Stress tests cover geometry degradation, periodic 1--5-s LiDAR absence, IMU
weighting, initialization error, and dynamic startup. A history-matched switch
separates fixing gravity from differences accumulated before an outage.
We report vertical position RMSE (RMSE$_z$) and 3D position RMSE (ATE), since
vertical improvement can accompany a worse trajectory. Absolute differences
help judge whether a large percentage change is practically important.

This work contributes:
\begin{itemize}
 \item within-system gravity and bias ablations in FAST-LIO2 and LIO-SAM,
 using actual state removal rather than suppressed corrections;
 \item tests of when state removal matters, including matched recovery after
 LiDAR outages and sensitivity to motion during initialization; and
 \item a direction-factor study showing that a smaller same-IMU residual
 does not guarantee lower vertical or 3D position error.
\end{itemize}

\noindent{\boldmath\textbf{Default: keep both $\mathbf g$ and $\mathbf b_a$
online. The tested same-IMU direction factor is not a generic $z$-drift remedy.}}

\section{Related Work}
\label{sec:related}

\textbf{State representation.}
Modern LIO includes iterated filters, smoothers, and continuous-time
estimators. FAST-LIO/FAST-LIO2 place gravity on $S^2$ inside an iterated
error-state filter \cite{xu2021fastlio,xu2022fastlio2,he2021ikfom}, and
Point-LIO retains the same convention \cite{he2023pointlio}. VE-LIOM also
estimates gravity online in an optimization framework \cite{gao2024veliom},
whereas LIO-SAM initializes a gravity-aligned frame without an equivalent
online direction state \cite{shan2020liosam}. Related pipelines also differ in
feature construction and map access. LOAM introduced edge/plane registration
\cite{zhang2014loam}, while FAST-LIO2 couples raw points to an incremental
$k$-d tree \cite{cai2021ikdtree}. Direct and continuous-time systems change
scan matching and propagation together
\cite{wang2023dliom,nguyen2023slict,chen2023dlio}. Cross-system accuracy
therefore confounds gravity handling with the front end, trajectory model,
map, and parameter settings. Our primary intervention changes only the compiled state
manifold; LIO-SAM provides a descriptive transfer check.

\textbf{Gravity--attitude--bias observability.}
Gravity direction, roll/pitch, and accelerometer bias have coupled
linearizations. Visual--inertial consistency analyses identify unobservable
and weakly observable directions \cite{hesch2014consistency,qin2018vinsmono},
and planar motion further limits excitation \cite{wu2017vinswheels}. LIO
initializers consequently estimate gravity, biases, timing, and extrinsics
before normal operation \cite{zhu2022liinit}; joint on-manifold calibration
makes the coupling explicit \cite{nemiroff2023gravity}. These analyses
establish when gravity is identifiable, but not whether its continued freedom
improves pose under frequent LiDAR correction. We test the latter while
intervening independently on $\mathbf g$ and $\mathbf b_a$. Mature
visual--inertial systems combine observability decisions with initialization,
relocalization, and map management \cite{campos2021orbslam3}; their aggregate
robustness cannot isolate one state's post-initialization utility.

\textbf{Gravity-enhanced residuals.}
Gravity-constrained registration removes rotational freedom using an
IMU-derived vertical direction \cite{kubelka2022gravity}. Recent radar--LiDAR
and radar--leg estimators add velocity-supported or soft $S^2$ gravity
information and report improved vertical accuracy
\cite{noh2025garlio,noh2025garlileo}. Elevator-specific models likewise show
that non-inertial motion can violate a nominal gravity model
\cite{elevator2026lio}. These methods add sensor information or residual
constraints; fixing an initialized gravity state does neither. We therefore
cross a same-IMU gravity-direction factor with fixed/online gravity and pair
its residual reduction with trajectory error. This design tests whether
tighter down-direction consistency reliably reduces vertical drift.

\textbf{Degradation and correction absence.}
Perceptual degeneracy is commonly detected from LiDAR geometry or estimator
observability and handled through selective updates, alternate odometry, or
direction-dependent weighting \cite{tagliabue2021lion,yao2025d2lio}. Weak
geometry and missing correction are often discussed within the same robustness
setting, although their estimator inputs differ. We separate them
experimentally: range/FoV degradation preserves the scan-update rhythm,
whereas periodic dropout removes correction entirely. Because the expected
effects are small, our protocol also audits reference-path plausibility and
exact correction timestamps, consistent with uncertainty-aware benchmark
generation \cite{hu2024paloc}. Our claim concerns an observation regime, not a
ranking of complete estimators. It distinguishes weak but accepted updates
from rejected or unavailable ones, a difference hidden by a single trajectory
score.

\section{Experimental Design}
\label{sec:method}

\subsection{State Interventions and Controls}
FAST-LIO2 is an iterated error-state filter on manifolds
\cite{xu2022fastlio2,he2021ikfom}. Its state comprises position $\mathbf p$,
orientation $\mathbf R$, LiDAR--IMU extrinsics $(\mathbf R_{LI},\mathbf t_{LI})$,
velocity $\mathbf v$, gyroscope bias $\mathbf b_g$, accelerometer bias
$\mathbf b_a$, and fixed-magnitude gravity $\mathbf g\in S^2$:
\begin{equation}
 d=3_{\mathbf p}+3_{\mathbf R}+6_{\mathrm{ext}}+3_{\mathbf v}
   +3_{\mathbf b_g}+3_{\mathbf b_a}+2_{\mathbf g}=23 .
 \label{eq:state_dimension}
\end{equation}
Online extrinsic calibration is disabled in every configuration. The
calibration remains necessary for deskewing and residual construction, but is
not an experimental variable. The disputed coupling enters propagation as
\begin{equation}
 \delta\dot{\mathbf v}\simeq\delta\mathbf g-\mathbf R\delta\mathbf b_a
 -\mathbf R[\mathbf a_m-\mathbf b_a]_{\times}\delta\boldsymbol\theta .
 \label{eq:coupling}
\end{equation}

State dimension is fixed at compile time, so four executables implement the
four configurations using otherwise identical code, front end, map,
calibration, initialization, and parameter settings. \methodA{} estimates
$\mathbf g$ and $\mathbf b_a$ online (23D); \methodG{} fixes $\mathbf g$
(21D); \methodB{} fixes $\mathbf b_a$ (20D); and \methodGB{} fixes both (18D).
Fixed quantities are removed from the estimated state and retained as
initialized constants in propagation. For comparison, a 23D proxy zeros their
corrections without changing the manifold. Because this proxy retains their
covariance blocks and influence on the Kalman gain, it tests the validity of
correction nulling, not state removal.

\begin{figure}[t]
\centering
\includegraphics[width=\columnwidth]{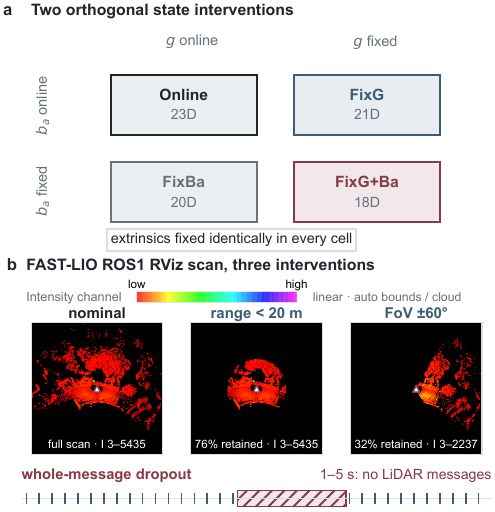}
\caption{Controlled design. \textbf{a}, Orthogonal compile-time state
interventions; extrinsics are held fixed identically, not removed.
\textbf{b}, One deskewed FAST-LIO scan, colored by retained point intensity
under nominal, 20-m range, and $\pm60^\circ$ FoV admission. The panels use the
ROS1 RViz \texttt{/cloud\_registered} intensity mapping with deterministic
display decimation; dropout removes all LiDAR messages during 1--5-s gaps.}
\label{fig:design}
\end{figure}

We repeat the state intervention in LIO-SAM \cite{shan2020liosam}. Native
\lsFGBA{} fixes gravity and estimates $\mathbf b_a$; \lsFGBZ{} also removes
$\mathbf b_a$ and its random walk. \lsGEBA{}/\lsGEBZ{} introduce a shared 2D,
fixed-magnitude online gravity direction, while $\mathbf b_g$ remains online.
The state dimension changes explicitly rather than through a near-zero prior.
The full ablation spans \NumLioSamSeq{} sequences and serves as a descriptive
cross-architecture check, not a second equivalence population.

\subsection{Correction Regimes and Direction Factor}
The FAST-LIO2 nominal population contains \NumCoreSeq{} public sequences from
MCD, TIERS, and M2DGR \cite{nguyen2024mcd,sier2023tiers,yin2022m2dgr}. They span
vehicle, handheld, and quadruped platforms with four LiDAR/IMU combinations;
path lengths range from \MinPath{} to \MaxPath{}. We modify one MCD vehicle
sequence offline by capping range at 20\,m, restricting horizontal field of
view (FoV) to a forward $\pm60^\circ$ sector, or deleting consecutive LiDAR messages over
1-, 2-, 3-, or 5-s intervals repeated every 20\,s.
The geometric cuts remove distant returns or lateral/rear coverage while
preserving scan timestamps. They are fixed geometric controls, not calibrated
models of sensor failure.
All four state configurations are run three times for the 2- and 3-s gaps;
a TUHH handheld trajectory provides a separate test of 2/3/5-s gaps.
Only the point-cloud stream changes. Paired configurations share the same IMU
data, calibration, reference trajectory, first/last timestamps, and outage
schedule. Here, \emph{weak geometry} means accepted updates with fewer spatial
constraints; \emph{correction absence} means no LiDAR update during an interval
(Fig.~\ref{fig:design}). The known dropout schedule isolates correction absence
without modeling front-end rejection or update delays.

The IMU-weight sweep pairs \methodA{}/\methodG{} and scales IMU noise and bias
random walk from \ImuNoiseMin{} to \ImuNoiseMax{} on \NumImuStressSeq{}
vehicle trajectories. A separate \NumInitStressRuns{}-run test injects
\InitStressAngles{} gravity-direction errors after shared initialization on one
vehicle and one handheld trajectory, then pairs \methodA{}/\methodG{}.
The startup experiment compares all four configurations on
60-s suffixes. Each trajectory contributes one quasi-static and
\NumDynamicInitPhases{} low-rate phases, supplemented by
\NumFastDynamicInitPairs{} stronger phases selected without viewing pose
outcomes. Within each phase, all configurations use the same bag offset.
A 150-ms IMU/ground-truth descriptor selects phases before pose evaluation. The resulting
\NumDynamicFactorialPhases{} phases and \NumDynamicFactorialRuns{} trajectories
test the shared acceleration-mean initializer, not a motion-aware alternative.

A \NumCoupledStateRuns{}-run, 60-s allocation test on one vehicle and one
handheld trajectory tilts $\mathbf g$ by $\pm\CoupledStateTiltDeg{}^\circ$
alone or pairs it with $\mathbf b_{a,1}=\mathbf b_{a,0}+\mathbf
R_0^\top(\mathbf g_1-\mathbf g_0)$, preserving $\mathbf g-\mathbf R_0\mathbf
b_a$ initially. Both states remain online; the test probes allocation
ambiguity, not absolute bias calibration.

A runtime audit alternates \methodA{}/\methodG{} over \NumRuntimePairs{} serial
pairs on MCD NTU Day10. After \NumRuntimeWarmupScans{} warm-up scans,
\NumRuntimeTimedScans{} matched scans per run provide core, matching, and
filter-algebra times; the last is cumulative update minus Jacobian construction.
Whole-core timing remains descriptive because map workloads bifurcate.

To control for pre-outage history, we fix gravity at its Online mean at a
pre-specified trigger and condition the retained covariance as
$P_{x\mid g}=P_{xx}-P_{xg}P_{gg}^{-1}P_{gx}$, then continue propagation and
measurement updates in a 21D error subspace. Unlike independently initialized
\methodG{}, this matched switch preserves the pre-switch state, map, and
posterior history; only uncertainty changes at the trigger. Clean/dropout
pairs share the trigger and are bit-identical before switching. Each
original-phase pair is run three times serially, and two further pre-specified
dropout starts test sensitivity to outage phase on each trajectory.
For either error metric $E$, the interaction is the switch-minus-Online
difference under dropout minus the same difference under clean input,
reported in meters.

To test the added residual, the LIO-SAM graph receives at each
post-initialization LiDAR correction epoch
\begin{equation}
 \mathbf r_{d,k}=\mathbf B(\mathbf d_k^{\mathrm{AHRS}})^\top
 \frac{\mathbf R_k^\top\hat{\mathbf g}}{\|\hat{\mathbf g}\|},
 \label{eq:direction_factor}
\end{equation}
where $\mathbf d_k^{\mathrm{AHRS}}$ is the body-frame down direction from the
latest attitude and heading reference system (AHRS) quaternion.
The normalized $\mathbf R_k^\top\hat{\mathbf g}$ is the direction predicted
from orientation and the fixed or estimated gravity direction. The columns of $\mathbf B$ form an
orthonormal tangent basis at $\mathbf d_k^{\mathrm{AHRS}}$.
This is GTSAM's two-component Unit3 direction residual, not an exact
spherical logarithm. Reported direction-residual norms scale $\|\mathbf r_{d,k}\|$ by
$180/\pi$ and are small-angle equivalents.
AHRS estimation and preintegration use measurements from the same IMU, so the
factor adds no independent sensor. It observes neither yaw, height, nor vertical
velocity. We compare factor-off with
$\sigma_d\in\{0.5^\circ,2^\circ,5^\circ\}$ and fixed/online gravity on two
continuous-correction sequences. Under 5-s gaps, Hall05 and TUHH Day04
(handheld) each receive the full $2\times3$ gravity-state-by-weight design,
with \NumDirRepeatsPerCell{} serial factor-on/off pairs for each state--weight
combination. We report each
trajectory separately. Factor covariance is a controlled weight, not a claim
that AHRS error is independent of preintegration.
Factor effects use the state- and repeat-matched off run from this campaign,
not the separate state-ablation baseline.
The factor belongs to LIO-SAM's incremental IMU--LiDAR estimation graph, not
the downstream loop-closure pose graph. Gravity priors in global pose-graph
optimization (PGO) are outside this intervention.

\subsection{Inference and Admission}
We report vertical position RMSE (RMSE$_z$) and 3D position RMSE (ATE) after
rigid $SE(3)$ alignment, without scale fitting. FAST-LIO2 fits a prefix that
spans both \AlignSec{} and \AlignM{}; LIO-SAM uses only the time criterion. The
protocol is fixed within each system and is not used to rank absolute errors
across systems. Tables abbreviate RMSE$_z$ as $z$. Percent change is used only
within a paired trajectory and is interpreted alongside absolute error. The
nominal \methodG{} comparison uses two one-sided tests (TOST) on paired log
ratios, with a pre-specified $\pm$\EquivMargin{} margin and 90\% confidence
intervals. We also report
paired differences in meters because ratios magnify small changes around
accurate baselines. A post-hoc $\pm$\StrictEquivMargin{} TOST and equal-weight
trajectory-family analysis are sensitivity checks, not replacement tests. A
paired Wilcoxon test evaluates
the factorial interaction
$I=\Delta_{g,b_a}-\Delta_g-\Delta_{b_a}$. LIO-SAM and mechanism sweeps remain
descriptive; no cross-trajectory pooled inference is made.

Admission requires six gates: plausible reference arc, matched attitude frame,
exact modal frame count, no host sleep, launch-specific binary fingerprint, and
serial execution. The early FAST-LIO2 logger hashed the main rather than each
reduced executable; state logs and build times support the intended structures,
but their exact hashes cannot be recovered. Later batches use launch-specific
fingerprints. Structural audits require removed states to remain bit-exact,
retained states to move, $\|\mathbf g\|=9.8090$\,m/s$^2$, and initial direction
agreement within $0.1^\circ$. LIO-SAM additionally checks correction times,
state dimension, factor count, AHRS age, and resets. A reset contributes a
stability outcome but no accuracy value. Complete finite trajectories outside
the estimated-arc diagnostic band remain as ``A'' warnings to avoid
outcome-conditioned admission. A Hall05 run with misaligned correction
timestamps was excluded and rerun; the final reset was not replaced.

For the matched APE traces, each pair shares the rigid transform fitted to its
Online control; the plotted center and band are the median and full range of
the three serial pairs. This preserves the zero pre-switch contrast and avoids
selecting a representative failure run.

\section{Results}
\label{sec:results}

\begin{figure}[t]
\centering
\includegraphics[width=\columnwidth]{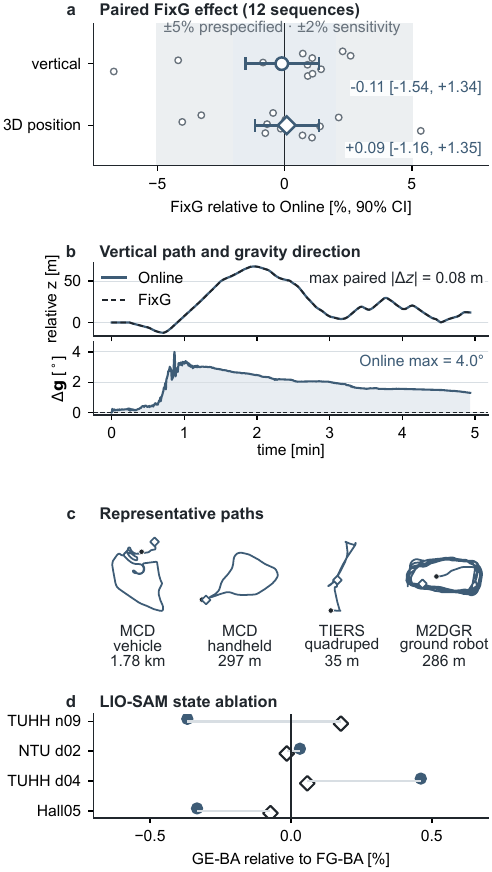}
\caption{Continuous-correction state ablation. \textbf{a}, FAST-LIO2
mean paired effects with 90\% confidence intervals ($n=\NumCoreSeq{}$ paired
sequences; small symbols show individual effects). \textbf{b}, A paired
sequence in which online gravity moves while Online and FixG vertical estimates
remain nearly superposed; this trace is illustrative, not the population test.
\textbf{c}, Representative \methodA{} trajectories; XY scales are independent,
and path lengths come from admitted references.
\textbf{d}, Descriptive LIO-SAM transfer; circles denote RMSE$_z$ and
diamonds ATE (3D position error).}
\label{fig:factorial}
\end{figure}

\subsection{Structural Validation}
Structural logs verify the intended intervention before accuracy is examined.
Online gravity moves by as much as \GravWanderMax{}, while every removed
gravity trace is bit-exact and retains the prescribed magnitude. Online
$\mathbf b_a$ changes and fixed $\mathbf b_a$ does not. Paired correction
timestamps, admitted frame counts, and initial directions match. The early
fingerprint limitation is stated in Sec.~\ref{sec:method}. These logs confirm
distinct state treatments, not an accuracy benefit. The FAST-LIO2
nominal block contains \NumCoreRuns{} admitted trajectories across the four
configurations. In LIO-SAM, exact state dimensions match the intended structures,
and every admitted factor run contributes one direction factor at each audited
epoch. The final Hall05 reset is retained as a stability outcome without an
accuracy value; the earlier timestamp-misaligned run was excluded and rerun.

\subsection{Continuous-Correction State Ablation}
Across the \NumCoreSeq{} nominal sequences, the signed median RMSE$_z$ change
for \methodG{} is \GZMedian{}, and the median absolute change is
\GZAbsMedian{} (range \GZRange{}). The corresponding ATE values are
\GATEMedian{} and \GATEAbsMedian{} (range \GATERange{}). TOST places both
mean paired effects inside the pre-specified $\pm$\EquivMargin{} margin
(both $p$ values \EquivPBoth{}). Back-transformed mean effects and 90\%
intervals are \EquivCIZ{} for RMSE$_z$ and \EquivCIATE{} for ATE
(Fig.~\ref{fig:factorial}; Table~\ref{tab:nominal}). This supports average
practical equivalence, not a per-trajectory guarantee. Both metrics also pass
a post-hoc $\pm$\StrictEquivMargin{} sensitivity test (maximum $p$
\StrictEquivPBoth{}), while equal weighting across five trajectory families
retains the pre-specified decision (maximum $p$ \FamilyEquivPBoth{}). A
launch-fingerprint reproduction on the same sequence units gives
\ReproEquivCIZ{}/\ReproEquivCIATE{} and passes both margins (maximum strict
$p$ \ReproStrictEquivPBoth{}); it is a reproducibility check, not an
independent $n=24$ population.
The paired trace in Fig.~\ref{fig:factorial}b shows a moving gravity state
alongside nearly superposed vertical estimates. The equivalence result comes
from the paired population test, not this individual trace.

Absolute scale matters: median absolute \methodG{}--\methodA{} differences
are \FixGAbsDeltaZMedian{} for RMSE$_z$ and \FixGAbsDeltaATEMedian{} for ATE.
On m2dgr-hall, \HallFixGZ{} is only \HallFixGZDeltaMm{} in RMSE$_z$; its ATE
change is \HallFixGATE{} (\HallFixGATEDeltaMm{}). We therefore interpret paired
percentages with absolute differences and require both metrics, which can
disagree in sign.

The LIO-SAM state ablation yields a similar nominal pattern. Across
\NumLioSamSeq{} complete sequences, \lsGEBA{} relative to native \lsFGBA{}
changes RMSE$_z$ by \mbox{\LioSamGZRange{}} and ATE by
\mbox{\LioSamGATERange{}}; neither metric has a consistent sign. Fixing both
quantities changes RMSE$_z$/ATE by
\mbox{\LioSamBothZRange{}}/\mbox{\LioSamBothATERange{}}. These data provide a
descriptive cross-architecture check on the absent nominal gain from online
gravity. They do not establish LIO-SAM equivalence or a general rule for
$\mathbf b_a$.

\begin{table}[t]
\caption{State ablation with continuous correction.}
\label{tab:nominal}
\centering
\footnotesize
\setlength{\tabcolsep}{2.5pt}
\begingroup
\renewcommand{\arraystretch}{1.06}
\newcommand{\nominalpair}[2]{\makebox[2.6em][r]{$#1$}\,$/$\,\makebox[2.6em][l]{$#2$}}
\begin{tabular*}{\columnwidth}{@{\extracolsep{\fill}}lcccc@{}}
\toprule
\rowcolor{TableBand}
\multicolumn{5}{@{}l}{\textsc{FAST-LIO2}\enspace paired change vs. \methodA{} [\%], $n=12$} \\
Configuration & dim. & median $z$/ATE & range $z$ & range ATE \\
\midrule
\methodG{} & 21 & \nominalpair{+0.9}{-0.1} & $[-6.7,+2.6]$ & $[-4.0,+5.4]$ \\
\methodB{} & 20 & \nominalpair{+1.5}{+0.2} & $[-2.1,+21.9]$ & $[-3.3,+12.6]$ \\
\methodGB{} & 18 & \nominalpair{+1.9}{+0.2} & $[-8.7,+47.2]$ & $[-6.1,+14.6]$ \\
\bottomrule
\end{tabular*}

\vspace{3pt}
\begin{tabular*}{\columnwidth}{@{\extracolsep{\fill}}lccc@{}}
\rowcolor{TableBand}
\multicolumn{4}{@{}l}{\textsc{LIO-SAM}\enspace paired $\Delta z/\Delta$ATE vs. \lsFGBA{} [\%]} \\
Sequence & \lsFGBZ{} & \lsGEBA{} & \lsGEBZ{} \\
\midrule
Hall05 & \nominalpair{+0.76}{+0.02} & \nominalpair{-0.33}{-0.07} & \nominalpair{+1.27}{+0.01} \\
TUHH d04 & \nominalpair{-3.20}{-1.09} & \nominalpair{+0.46}{+0.06} & \nominalpair{+0.93}{+0.16} \\
NTU d02 & \nominalpair{+0.04}{+0.02} & \nominalpair{+0.03}{-0.01} & \nominalpair{+0.00}{-0.04} \\
TUHH n09 & \nominalpair{-0.22}{+0.27} & \nominalpair{-0.37}{+0.18} & \nominalpair{-0.09}{+0.46} \\
\bottomrule
\end{tabular*}
\endgroup

\end{table}

\begin{figure}[t]
\centering
\includegraphics[width=\columnwidth]{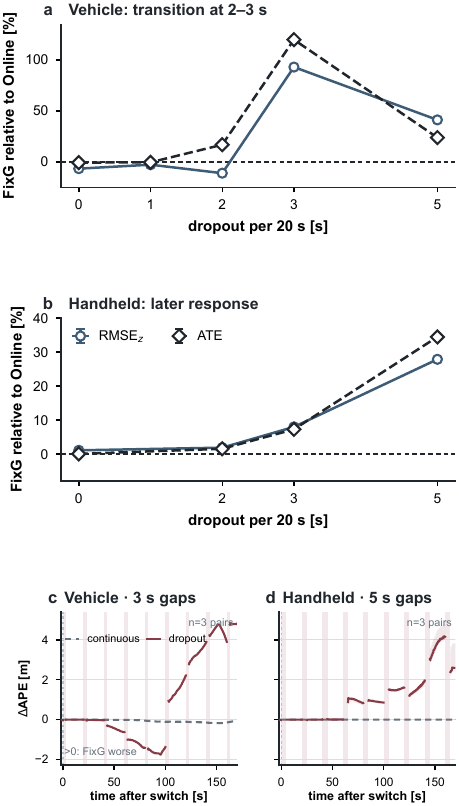}
\caption{Correction-absence boundary. \textbf{a,b}, \methodG{} relative to
\methodA{} on vehicle and handheld trajectories; both RMSE$_z$ and ATE are
required. \textbf{c,d}, APE difference after the matched 23D-to-21D switch:
median and range across \NumWarmRepeatPairs{} serial pairs under a common
alignment. Shading denotes missing corrections; positive values favor Online.}
\label{fig:boundary}
\end{figure}

\subsection{Weak Geometry and Correction Absence}
Continuously degraded scans do not reproduce a dropout penalty. A 20\,m range
cap changes \methodG{} RMSE$_z$/ATE by \RangeGZ{}/\RangeGATE{}; a forward
$\pm60^\circ$ FoV changes them by \FovGZ{}/\FovGATE{}. Both retain scan-update
timing (Table~\ref{tab:operating}).

Complete dropout separates the designs. A 1-s gap every 20\,s remains benign.
At 2\,s, vehicle RMSE$_z$/ATE move in opposite directions
(\DropTwoGZ{}/\DropTwoGATE{}), so neither metric alone supports a configuration
choice. At 3\,s, both rise by \DropThreeGZ{}/\DropThreeGATE{} in all
\NumBoundaryRepeats{} serial repeats. The handheld response is later:
\HhsDropThreeGZ{}/\HhsDropThreeGATE{} at 3\,s and
\HhsDropFiveGZ{}/\HhsDropFiveGATE{} at 5\,s. Vehicle 5-s effects remain worse
than Online but are smaller than at 3\,s, so duration is not a monotonic dose.

The matched APE traces in Fig.~\ref{fig:boundary}c,d start at zero, stay near
zero with continuous correction, but separate progressively after repeated
gaps. The vehicle contrast initially changes sign; the handheld responds later.
The repeated trajectories do not show an effect confined to the first return.
The 2--3-s vehicle transition and later handheld response place the onset at
different gap durations, rather than a universal threshold.

The matched switch reproduces the regime contrast across
\NumWarmRepeatPairs{} serial pairs at the original dropout phase. Clean-input
ATE effects are small: \mbox{\WarmVehCleanATE{}} on the vehicle and
\mbox{\WarmHhsCleanATE{}} on the handheld trajectory. The median dropout-minus-clean
ATE interactions are \mbox{\WarmVehATEInteraction{}} and
\mbox{\WarmHhsATEInteraction{}}, respectively. Across \NumWarmPhases{}
dropout phases, ATE interactions are positive on both trajectories, whereas
RMSE$_z$ interactions cross zero (Table~\ref{tab:operating}). The stable effect
across these phases is therefore in 3D position error, not vertical error.

\begin{table}[t]
\caption{Correction-regime boundary.}
\label{tab:operating}
\centering
\footnotesize
\setlength{\tabcolsep}{2.8pt}
\begin{tabular}{@{}lcccc@{}}
\toprule
Condition & \methodA{} $z$/ATE [m] & $\Delta z$ [\%] & $\Delta$ATE [\%] & $n$ \\
\midrule
\rowcolor{TableBand}
\multicolumn{5}{@{}l}{\textsc{Vehicle}\enspace continuous correction} \\
nominal & $2.49/6.60$ & $-6.7$ & $-0.8$ & 1 \\
range 20 m & $2.81/10.08$ & $+0.2$ & $-4.0$ & 1 \\
FoV $\pm60^\circ$ & $5.17/6.49$ & $+6.9$ & $+3.5$ & 1 \\
\addlinespace[2pt]
\rowcolor{TableBand}
\multicolumn{5}{@{}l}{\textsc{Vehicle}\enspace dropout per 20 s} \\
1 s/20 s & $2.43/6.50$ & $-2.8$ & $-0.5$ & 1 \\
2 s/20 s & $2.31/6.27$ & $-11.2$ & $+16.8$ & 3 \\
3 s/20 s & $3.09/8.42$ & $+93.1$ & $+119.9$ & 3 \\
5 s/20 s & $4.92/13.65$ & $+41.2$ & $+23.8$ & 1 \\
\addlinespace[2pt]
\rowcolor{TableBand}
\multicolumn{5}{@{}l}{\textsc{Handheld}\enspace dropout per 20 s} \\
2 s/20 s & $0.445/0.519$ & $+1.9$ & $+1.5$ & 1 \\
3 s/20 s & $0.485/0.536$ & $+8.0$ & $+7.3$ & 1 \\
5 s/20 s & $3.28/4.07$ & $+27.9$ & $+34.5$ & 1 \\
\addlinespace[2pt]
\rowcolor{TableBand}
\multicolumn{5}{@{}l}{\textsc{Matched posterior switch}\enspace dropout--clean interaction [m]} \\
vehicle repeats & -- & $+0.62$ & $+4.16$ & 3 \\
vehicle phases & -- & $[-0.23,+2.74]$ & $[+0.87,+4.16]$ & 3 \\
handheld repeats & -- & $+0.09$ & $+1.01$ & 3 \\
handheld phases & -- & $[-0.08,+0.22]$ & $[+0.30,+1.09]$ & 3 \\
\bottomrule
\end{tabular}

\end{table}

\begin{figure}[t]
\centering
\includegraphics[width=\columnwidth]{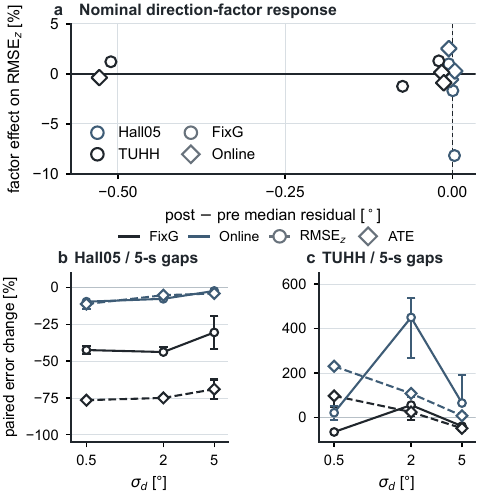}
\caption{Direction-factor response. \textbf{a}, Nominal residual-norm
(degree-equivalent) and RMSE$_z$ changes; color denotes trajectory, shape
denotes gravity state.
\textbf{b,c}, Dropout effects: paired medians and full ranges, with lines
connecting tested weights. Positive means worse. Failures and warnings
remain in Table~\ref{tab:direction}.}
\label{fig:direction}
\end{figure}

\subsection{Direction-Factor Response}
Nominally, factor strength has no monotonic dose response
(Fig.~\ref{fig:direction}; Table~\ref{tab:direction}). With fixed gravity, the
tightest factor worsens RMSE$_z$ by \DirStrongFGZRange{}, whereas the
intermediate factor changes RMSE$_z$/ATE by
\DirMidFGZRange{}/\DirMidFGATERange{}. Online-gravity ATE stays within
\DirGEATERange{}. Across these configurations, smaller direction residuals do not
predict smaller trajectory errors.

Under repeated 5-s gaps, the two trajectories respond differently. On Hall05, all
\NumHallDirBeneficialCells{} state--weight medians improve both errors, but the
FixG/$5^\circ$ configuration yields only $2/3$ accuracy runs because one planned run
resets; that failure is counted and was not replaced. TUHH contains both sign
reversal and metric disagreement. With FixG/$0.5^\circ$, RMSE$_z$ changes by
\TuhhDirFGStrongZ{} while ATE changes by \TuhhDirFGStrongATE{}. With online
gravity/$2^\circ$, both deteriorate by \TuhhDirGEZTwo{}/\TuhhDirGEATETwo{}.
The \NumTuhhDirArcWarnings{} complete TUHH trajectories outside the estimator
arc diagnostic band retain their metrics and carry ``A'' warnings; neither was
replaced. Thus a lower vertical error can conceal worse 3D pose, and a direction
factor can worsen both metrics. Neither increasing factor strength nor keeping
gravity online ensures an improvement.

\subsection{Accelerometer Bias and Gravity--Bias Coupling}
Accelerometer bias does not inherit gravity's removal result. \methodB{} and
\methodGB{} have median RMSE$_z$ changes of \BAMedianZ{} and
\BothMedianZ{}, with larger sequence-level ranges than \methodG{}
(Table~\ref{tab:nominal}). True-manifold interaction is \InteractionZ{}
($p=$\InteractionPZ{}) for RMSE$_z$ and \InteractionATE{}
($p=$\InteractionPATE{}) for ATE, providing no evidence of superadditivity.
The correction-nulling proxy biases the fixed-$\mathbf b_a$
comparison by \ProxyBAZ{}, confirming that it is not true removal.

The paired intervention shifts $\mathbf b_a$ by \CoupledStateInitialBA{} while
preserving $\mathbf g-\mathbf R\mathbf b_a$ within
\CoupledStateInitialHMax{}. In the handheld final 10\,s, gravity/bias remain
\CoupledHhsGRange{}/\CoupledHhsBARange{} apart, but the effective term/position
differ by only \CoupledHhsHRange{}/\CoupledHhsPoseRange{}; the vehicle estimates
return close to baseline. Handheld poses thus remain similar despite distinct
gravity and bias estimates.

\subsection{IMU Weight and Initialization}
IMU weighting does not reveal a uniform advantage for either gravity choice.
On vehicle NTU Day10, \methodG{} changes RMSE$_z$ by
\ImuDayZMin{} to \ImuDayZMax{} and ATE by \ImuDayATEMin{} to
\ImuDayATEMax{}. On vehicle NTU Night04, RMSE$_z$ changes by
\ImuNightZMin{} to \ImuNightZMax{}, but ATE worsens by
\ImuNightATEMin{} to \ImuNightATEMax{}. Maximum roll/pitch change is
\ImuStressAttMax{}.

The initialization-direction stress test also shows no monotonic FixG penalty.
Across \NumInitStressRuns{} paired runs at \InitStressAngles{}, RMSE$_z$/ATE
changes span \InitErrZMin{} to \InitErrZMax{}/\InitErrATEMin{} to
\InitErrATEMax{}, and the maximum roll/pitch change is \InitErrAttMax{}.

Across \NumDynamicFactorialPhases{} start phases and
\NumDynamicFactorialRuns{} trajectories, the dynamic-start $2\times2$ yields
no universal gravity choice. At one
stronger vehicle phase, Online and FixG produce
\FastDynamicRescueOnlineATE{} and \FastDynamicRescueFixGATE{} ATE,
respectively, while RMSE$_z$ reverses
(\FastDynamicRescueOnlineZ{} versus \FastDynamicRescueFixGZ{}). Fixed bias is
more sensitive to startup phase: both fixed-bias configurations diverge at one
low-rate handheld start. FixBa also worsens ATE in all
\NumStrongFixBaATEWorse{} stronger-motion phases (\StrongFixBaATERange{}),
although it helps at other low-rate phases. The
\NumDynamicFixedBaFailureOutcomes{} complete, finite failure outcomes are
retained rather than rerun. The phase- and metric-dependent signs do not yield
a uniform vertical Online gain.

\section{Discussion}
\label{sec:analysis}

\subsection{Why Scan Recovery Changes the Comparison}
Accepted LiDAR scans repeatedly constrain pose, so gravity-state drift need
not produce a worse trajectory. During an outage, errors accumulate until
registration resumes from the propagated state.

For a small gravity-direction discrepancy $\varepsilon$ propagated without an
external update for time $T$, the leading position contribution is
\begin{equation}
 \|\delta\mathbf p_g\|\approx\tfrac12g\sin(\varepsilon)T^2 .
 \label{eq:timescale}
\end{equation}
This describes displacement before recovery, not final trajectory error.
Registration may correct a large error but leave a smaller one partly
uncorrected, depending on the scene and returning pose. This is consistent
with the non-monotonic 3-s and 5-s results, although we do not directly measure
the registration convergence basin.

In the matched switch, both runs start from the same state and map, but fixing
gravity conditions the covariance. Gravity has no process dynamics, so the
nominal means remain equal during the first gap; Eq.~\ref{eq:timescale} cannot
explain separation before scan return. Differences arise during correction,
through the changed covariance and continued Online gravity updates. The APE
traces show accumulation over successive recovery corrections. ATE
interactions stay positive across tested starts, while RMSE$_z$ interactions
change sign. The repeated advantage concerns 3D recovery, not direct height
information.

The relevant interval $T_k=t_k-t_{k-1}$ is between \emph{accepted pose
corrections}, not arriving LiDAR packets. A rejected scan cannot interrupt
inertial error accumulation. Message removal controls these intervals, whereas
natural rejection may depend on geometry and the estimated pose. Biased but
accepted registrations instead introduce erroneous corrections, a different
failure mode not tested here. Scan rate alone is therefore insufficient to
transfer the observed thresholds.

\subsection{Interpreting the Direction-Factor Response}
The factor uses AHRS directions derived from the IMU measurements also used
for preintegration. It observes neither height nor vertical velocity, so its
vertical-error effects are mediated by attitude, bias, propagation, and
registration.

At fixed gravity magnitude, a direction error $\varepsilon$ gives
\begin{equation}
 \begin{aligned}
  \|\mathbf e_g^\top\delta\mathbf g\|
    &=g(1-\cos\varepsilon)=\mathcal O(\varepsilon^2),\\
  \|\mathbf P_{\perp}\delta\mathbf g\|
    &=g\sin\varepsilon=\mathcal O(\varepsilon).
 \end{aligned}
\label{eq:gravity_components}
\end{equation}
where $\mathbf e_g$ is unit true down and
$\mathbf P_\perp=\mathbf I-\mathbf e_g\mathbf e_g^\top$.
Small direction errors therefore act to first order transversely, but only to
second order along true down. A direction factor can still change RMSE$_z$ by
rotating measured specific force or changing scan-matching initialization.
Neither route measures height, and Eq.~\ref{eq:gravity_components} alone does
not determine the final error change.

Hall05's improvements and TUHH's mixed or adverse responses reject an
unconditional benefit from stronger direction consistency. They do not
invalidate independent gravity, velocity, contact, radar, or registration
information
\cite{kubelka2022gravity,noh2025garlio,noh2025garlileo}, nor do they test a
downstream global pose-graph prior. Factor covariance must be validated with
the state structure, correction regime, and both trajectory metrics.

Unmodeled same-IMU correlation could contribute to the adverse response, but
we do not isolate it from weighting and online-gravity interactions.
A smaller residual establishes agreement with AHRS, not an independent
reference (Fig.~\ref{fig:direction}). Table~\ref{tab:direction} also retains
resets and arc warnings; reporting only successful accuracy comparisons would
overstate reliability.

\begin{table}[t]
\caption{Direction-factor effects relative to factor-off.}
\label{tab:direction}
\centering
\footnotesize
\setlength{\tabcolsep}{2.0pt}
\begingroup
\renewcommand{\arraystretch}{1.05}
\begin{tabular*}{\columnwidth}{@{\extracolsep{\fill}}llrrl@{}}
\toprule
Weight & $g$ state & $\Delta$RMSE$_z$ [\%] & $\Delta$ATE [\%] & outcome \\
\midrule
\rowcolor{TableBand}
\multicolumn{5}{@{}l}{\textsc{Hall05}\enspace 5-s dropout per 20 s} \\
$0.5^\circ$ & \methodG{} & $-42.4$ & $-76.5$ & 3/3 \\
$2^\circ$ & \methodG{} & $-43.8$ & $-75.0$ & 3/3 \\
$5^\circ$ & \methodG{} & $-30.5$ & $-69.1$ & 2/3; 1 R \\
$0.5^\circ$ & \methodA{} & $-9.6$ & $-11.2$ & 3/3 \\
$2^\circ$ & \methodA{} & $-7.6$ & $-5.2$ & 3/3 \\
$5^\circ$ & \methodA{} & $-2.3$ & $-4.0$ & 3/3 \\
\addlinespace[2pt]
\rowcolor{TableBand}
\multicolumn{5}{@{}l}{\textsc{TUHH handheld}\enspace 5-s dropout per 20 s} \\
$0.5^\circ$ & \methodG{} & $-64.6$ & $+97.6$ & 3/3; 1 A \\
$2^\circ$ & \methodG{} & $+56.2$ & $+25.0$ & 3/3 \\
$5^\circ$ & \methodG{} & $-38.3$ & $-48.1$ & 3/3 \\
$0.5^\circ$ & \methodA{} & $+21.8$ & $+230.9$ & 3/3; 1 A \\
$2^\circ$ & \methodA{} & $+450.0$ & $+108.4$ & 3/3 \\
$5^\circ$ & \methodA{} & $+65.4$ & $+8.2$ & 3/3 \\
\bottomrule
\end{tabular*}
\endgroup

\vspace{1pt}
{\scriptsize\raggedright Outcome entries report accuracy runs/planned runs;
R denotes a graph reset and A a retained arc warning.\par}
\end{table}

\subsection{Choosing a Default Configuration}
Keeping $\mathbf g$ and $\mathbf b_a$ online allows adjustment but does not
guarantee physical accuracy. The handheld runs retain different gravity and
bias estimates despite similar effective accelerations and poses. Vehicle
excitation largely removes those differences. As Eq.~\ref{eq:coupling}
suggests, accurate pose does not establish that gravity and bias have each
been identified.

With reliable initialization and continuous correction, FAST-LIO2 FixG is
equivalent on average but saves negligible compute. Keeping both states online
allows adaptation when motion contaminates startup or corrections disappear.
Fixed-bias variants are phase-sensitive, and one FixG start incurs large ATE
harm. We therefore recommend keeping both states online by default. FixG is an
option for a validated steady-state setting, not a generally better estimator.
A separate direction factor needs its own paired RMSE$_z$ and ATE validation
under the intended operating conditions.

Across \NumRuntimePairs{} serial pairs, FixG reduced filter algebra by
\RuntimeAlgebraReduction{}, but it occupied only \RuntimeAlgebraCoreShare{} of
core time versus \RuntimeMatchingCoreShare{} for scan matching. Its attributable
saving was \RuntimeAttributableCoreSaving{}, and whole-core change
(\RuntimeCoreMedianChangeRange{}) crossed zero. Dimension reduction is not a
runtime strategy.

Two dropout and two initialization trajectories cannot establish universal
thresholds. The factor study retains \NumHallDirResetOutcomes{} reset and
\NumTuhhDirArcWarnings{} arc warnings. Larger initialization errors,
calibration faults, and motion-aware initialization also remain untested.
TOST concerns pose error, not covariance
consistency; dropout results are not pooled across trajectories.

\vfuzz=3pt
\balance

\section{Conclusion}
\label{sec:conclusion}

With continuous LiDAR correction, FAST-LIO2's mean paired effects of fixing gravity
on vertical and 3D position error satisfy practical equivalence across
\NumCoreSeq{} dataset sequences. A descriptive
\NumLioSamSeq{}-sequence LIO-SAM ablation also
finds no consistent benefit from online gravity estimation. This does not
justify fixing gravity or bias by default. The
matched-dropout and dynamic-start tests show recovery and failure costs that
nominal averages miss, while the runtime saving is negligible. We recommend
keeping both $\mathbf g$ and $\mathbf b_a$ online, without treating their
individual estimates as calibrated measurements. FixG remains an optional,
validated steady-state simplification.
Do not add the tested same-IMU gravity-direction factor as a generic $z$-drift
remedy; the factor can reduce its residual while improving, leaving unchanged,
or worsening trajectory error. Independent gravity sensing and global loop-closure PGO priors remain
outside this claim.

\bibliographystyle{IEEEtran}
\bibliography{refs}

\end{document}